\documentclass[a4paper]{svproc}          %

\usepackage{url}

\usepackage{amsmath}
\usepackage{amssymb}
\usepackage{mathtools}
\usepackage{siunitx}            %

\usepackage{graphicx}           %
\usepackage{subcaption}         %
\usepackage{diagbox}            %
\usepackage{wrapfig}            %
\usepackage{stfloats}           %

\usepackage{xcolor}
\usepackage{xparse}             %

\renewcommand{\vec}[1]{\mathbf{#1}}          %
\newcommand{\mat}[1]{\mathbf{#1}}            %
\newcommand{\trans}{^\mathsf{T}}             %
\newcommand{\inv}{^{-1}}                     %

\newcommand{\R}{\mathbb{R}}                  %

\newcommand{\abs}[1]{\left|#1\right|}
\newcommand{\norm}[1]{\left\lVert#1\right\rVert}

\newcommand{\dd}{\,\mathrm{d}}

\newcommand{\pderiv}[2]{\frac{\partial #1}{\partial #2}}

\newcommand\scaleddot{\scalebox{.89}{.}}   %
\makeatletter
\newcommand{\dddot@fac}[1]{%
  \ifx#1\scriptscriptstyle 0.1775\else
  \ifx#1\scriptstyle       0.1878\else
                           0.1933\fi\fi}
\newcommand{\dddot@quad}[1]{%
  \ifx#1\scriptscriptstyle \fontdimen6\scriptscriptfont0\else
  \ifx#1\scriptstyle       \fontdimen6\scriptfont0\else
                           \fontdimen6\textfont0\fi\fi}
\newcommand{\dddot@}[2]{%
  \begingroup
  \edef\dddot@fam{\the\fam}%
  \def\dddot@sym{\fam\dddot@fam #2}%
  \setbox\@tempboxa\hbox{$\m@th#1\dddot@sym$}\@tempdimb\wd\@tempboxa
  \setbox\@tempboxa\hbox{$\m@th#1\dddot@sym\mathchar"17F$}\@tempdima\wd\@tempboxa
  \advance\@tempdima-\@tempdimb
  \setbox\@tempboxa\hbox{$\m@th#1\mathchar"17F$}\advance\@tempdima-\wd\@tempboxa
  \@tempdimb\dddot@fac{#1}\dddot@quad{#1}%
  \rlap{\kern\@tempdima\kern-\@tempdimb$\m@th#1\fam\dddot@fam\dot{\phantom{\dddot@sym}}$}%
  \rlap{\kern\@tempdima               $\m@th#1\fam\dddot@fam\dot{\phantom{\dddot@sym}}$}%
  \rlap{\kern\@tempdima\kern \@tempdimb$\m@th#1\fam\dddot@fam\dot{\phantom{\dddot@sym}}$}%
  #2%
  \endgroup}
\providecommand{\dddot}{}
\DeclareRobustCommand{\dddot}[1]{\mathord{\mathpalette\dddot@{#1}}}
\DeclareRobustCommand{\dddotbf}[1]{\dddot{\mathbf{#1}}}
\renewcommand{\ddddot}[1]{%
  {\mathop{\kern\z@#1}\limits^{\makebox[0pt][c]{\vbox to-2.2\ex@{\kern-\tw@\ex@
  \hbox{\normalfont\scaleddot\kern-0.5pt\scaleddot\kern-0.5pt\scaleddot\kern-0.5pt\scaleddot}\vss}}}}}
\makeatother

\newcommand{\trace}{\operatorname{tr}}

\newcommand{\argmin}{\operatorname*{arg\,min}}

\newcommand{\E}{\mathbb{E}}

\providecommand{\st}{}\renewcommand{\st}{\text{s.t.}}

\newcommand{\zeros}{\mathbf{0}}

\newcommand{\eye}{\mat{I}}

\newcommand{\q}{\mathbf{q}}
\newcommand{\dq}{\dot{\mathbf{q}}}
\newcommand{\ddq}{\ddot{\mathbf{q}}}
\newcommand{\dddq}{\dddotbf{q}}

\newcommand{\x}{\mathbf{x}}
\newcommand{\dx}{\dot{\mathbf{x}}}
\newcommand{\ddx}{\ddot{\mathbf{x}}}

\newcommand{\ball}{\mathbf{b}}

\newcommand{\g}{\mathbf{g}}

\newcommand{\emap}{\boldsymbol{\phi}}        %

\usepackage{tikz}
\usetikzlibrary{arrows.meta, positioning, fit, backgrounds, calc, shapes.geometric}
\usepackage{standalone}

\usepackage[hidelinks]{hyperref}   %
\usepackage[capitalize]{cleveref}

\newif\ifshowideas
\showideastrue

\begin{document}
\mainmatter

\title{Task-Error Residual Learning for Real-Robot Five-Ball Juggling}
\titlerunning{Task-Error Residual Learning for Juggling}

\author{Kai Ploeger\inst{1} \and Alap Kshirsagar\inst{1,4} \and Jan Peters\inst{1,2,3}}
\authorrunning{K.\ Ploeger, A.\ Kshirsagar and J.\ Peters}
\tocauthor{Kai Ploeger, Alap Kshirsagar, Jan Peters}

\institute{Technical University of Darmstadt, Germany
\and
German Research Center for AI (DFKI), Germany
\and
Hessian Center for Artificial Intelligence (hessian.AI), Germany
\and
Indian Institute of Technology Delhi -- Abu Dhabi, UAE\\[6pt]
\email{\{kai,alap,jan\}@robot-learning.de}}

\maketitle

\begin{abstract}

For residual learning that refines existing behavior, sample efficiency depends on two things: how much information each rollout returns, and how efficiently the learner uses that information.
Reinforcement learning's standard scalar reward carries far less information than the directional task error that defines the task.
Random exploration further discards whatever information each rollout returns.
Through residual learning with directional task-error supervision and a task error model that drives sample selection, we achieve stable three-, four-, and five-ball juggling on anthropomorphic Barrett WAM arms.
Despite planning and controlling through a simple, idealized stack, the system converges from the second attempt.
The first attempt drops, after which task error decreases monotonically without further failures.
In comparison, five-ball juggling typically takes humans years of practice.
We compare residual learners across two ternary axes, the directional information in the learning feedback and the commitment of the analytic prior, spanning Newton-style Jacobian updates, Composite Bayesian Optimization, and stochastic search methods.
Both axes prove necessary for fast convergence: neither directional feedback nor an informative prior alone matches their combination, and the simplest method that combines them, a fixed-Jacobian Newton update, is the fastest and fully reliable.
The learned residual tolerates substantial prior misalignment and degraded joint tracking, affecting mainly convergence speed.
The bottleneck for residual learning on real robots is therefore the information content of the supervision signal and how the learner uses it, not the accuracy of the surrounding stack.
Video documentation of all experiments is available at \url{https://kai-ploeger.com/residual-juggling}.
\end{abstract}

\clearpage
\section{Introduction}
\label{sec:introduction}

\begin{wrapfigure}{r}{0.45\textwidth}
    \vspace{-\intextsep}
    \centering
    \includegraphics[width=0.45\textwidth]{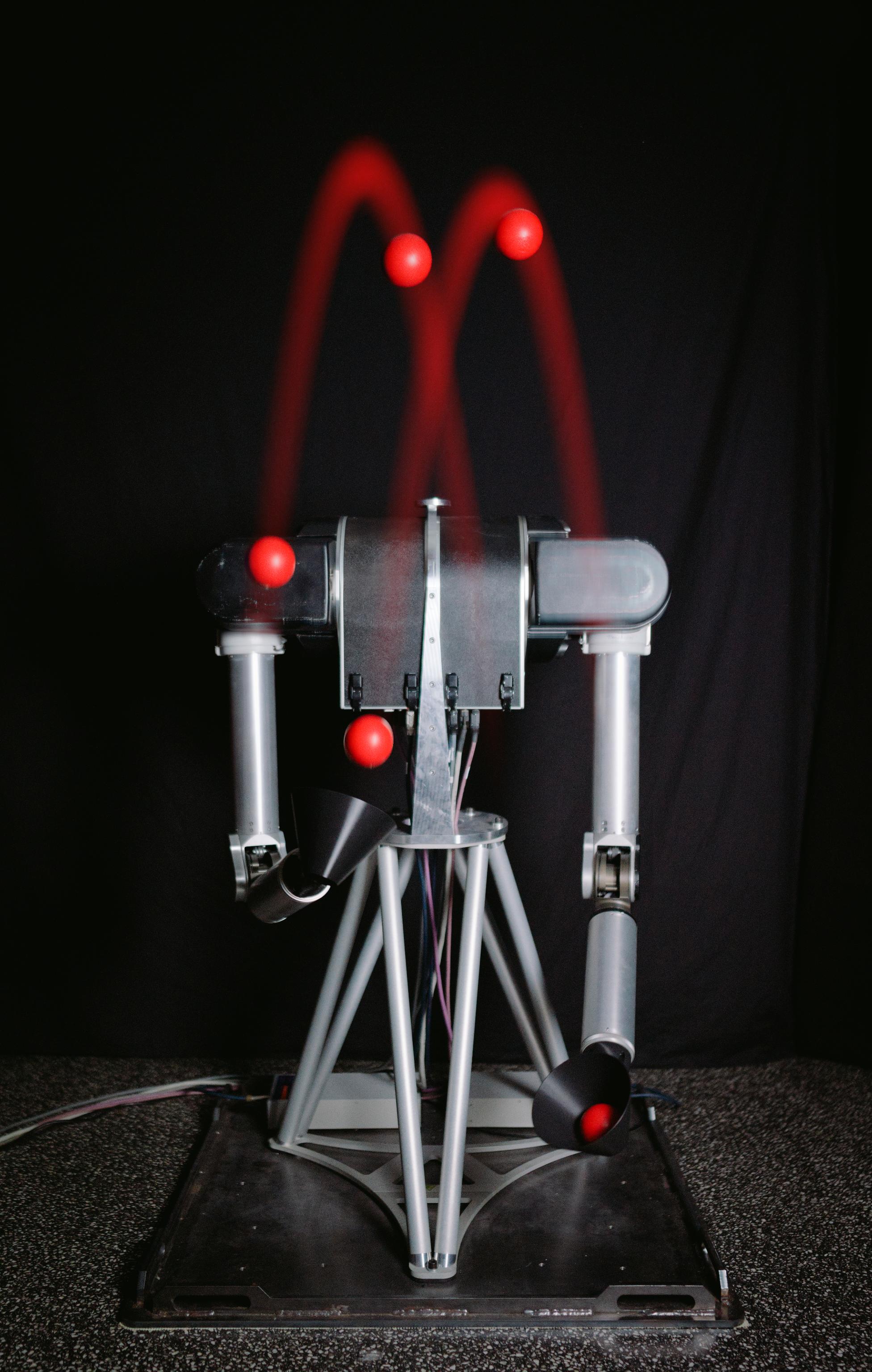}
    \caption{Two 4-DOF Barrett WAM arms juggling a uniform five-ball cascade. The residual learner reaches this pattern from the second attempt.}
    \label{fig:hero}
    \vspace{-0.5\intextsep}
\end{wrapfigure}
With task-error supervision, the rest of the robot-learning stack can be deliberately simple.
We make this case in the real-robot setting of five-ball juggling on two arms, executed open-loop and without simulation.
The task is unforgiving: fast, accurate throws on a tight schedule, and one miss ends the run.
The system, shown in \cref{fig:hero}, juggles reliably within a handful of real-robot attempts, despite kinematic planning through an idealized contact-switch model and a parabolic ballistic flight model, and tracking with a soft proportional-derivative (PD) controller and rigid-body feedforward.
Three design choices make this work: directional task-error supervision, a stack built for repeatability rather than accuracy, and model-driven exploration.

\paragraph{Directional Feedback.}
A scalar objective, the rewards or costs typical of reinforcement learning, collapses an inherently directional task error onto a single number, turning residual learning into local minimization rather than the easier problem of local root-finding.
This cost bites hardest for on-robot adaptation, where every sample is a real attempt.
Sim-based scalar reinforcement learning (RL) faces a different bottleneck, the sim-to-real gap for highly dynamic tasks.
Iterative learning control (ILC) already efficiently iterates root-finding on a signed tracking error, but at the joint-tracking level.
Its error lives between executed and reference joint trajectories rather than between intended and realized task outcomes, leaving it blind to environment interactions like contact dynamics.
Lifting root-finding to the task level directly targets the signal that defines task success, leading to fast convergence to optimal performance.
Measuring the displacement between intended and observed ball trajectories as the task error,
we observe that directional learners converge both faster and to lower final task error than scalar learners, which receive only its norm.

\paragraph{Stack Repeatability.}
MPC and related optimization\nobreakdash-based approaches demand accuracy of the dynamics model.
The broader assumption is that each layer of the stack should be made as accurate as possible.
The residual-learning alternative demands only repeatability, turning the unlearned stack into a fixed function the residual can correct around.
At the extremes, stack accuracy makes learning unnecessary, while stack repeatability makes it possible.
In between, stack accuracy controls how fast the learner converges and what the residual must compensate for.
The final task performance is the same, as long as the movement remains coupled to the command.
For fast movements, high control gains are conventionally chosen to reduce trajectory tracking error.
We show that trading stack accuracy for repeatability removes this demand, permitting softer gains.
Lower gains reduce the impact force in any unintended contact, a safety benefit for a highly dynamic task performed close to humans.
In our system, the planner uses a 1g contact-switch model and a parabolic ballistic predictor.
The controller tracks with soft PD gains and a rigid-body feedforward.
The residual learner adapts the takeoff velocity from offline-computed task-error labels, starting with no initial correction.
Although no part of the stack is accurate, its repeatability is enough for the system to juggle open-loop, with no active catching or in-loop perception.
Varying stack accuracy through control gains does not change final performance, only convergence speed.

\paragraph{Model-Based Exploration.}
Evolutionary search and reinforcement learning explore by random perturbation: generate samples, evaluate, update.
We instead build a task model of action-to-outcome and use it to explore directly, picking the next sample from the model, not from a noise distribution.
The form of the task model depends on the feedback.
The model is a local Jacobian for directional task error, or a scalar surrogate, such as a local quadratic, for scalar feedback.
In our comparison, model-driven exploration converges in fewer attempts than random exploration in both feedback regimes.
The combination of directional feedback with an informative prior is the most sample-efficient.
With directional task-error feedback the Jacobian is especially cheap to fit.
A few rollouts suffice, resulting in faster convergence than surrogate fitting under scalar feedback.
A task model is a convenient way to introduce inductive biases (priors, domain knowledge) into exploration, further increasing convergence speed, while random perturbation has nowhere to encode them.
Our inductive biases come from the idealized ballistic map: its Jacobian is a free analytical sensitivity prior for our learners.
We quantify the contribution of each of these choices in our method comparison (\cref{sec:exp:method-comparison}).

\medskip
We present three main contributions.
First, three-, four-, and five-ball juggling patterns on two anthropomorphic 4-DOF arms, converging by the second attempt across all three patterns under a simple stack with no simulation, no in-loop perception, and no active catching.
Second, an empirical map of a $3\times 3$ method matrix spanning learning feedback (directional, norm, squared-norm) and prior specificity (none, structural, calibrated), identifying which design choices reduce sample complexity.
Third, real-robot ablations quantifying the learner's tolerance to both prior misspecification (rotation of the analytic Jacobian) and stack inaccuracy (PD-gain sweep), showing that residual learning remains effective across a wide operating range.

\clearpage
\section{Related Work}
\label{sec:related-work}

Most robotic juggling work focuses on batting~\cite{schaalOpenLoopStable1993,rizziFurtherProgressRobot1994} or rolling~\cite{tanakaLearningRoboticContact2021,woodruffRoboticContactJuggling2023}.
Toss juggling releases balls into ballistic flight between contacts, demanding high end-effector velocities and forgoing the elastic-collision energy preservation available in batting.
It can be scaled to arbitrary difficulty, making it a strong platform for studying dynamic motor control and predictive modeling.

This paper builds on a sequence of our own prior work on juggling.
Two-ball juggling on a single arm was learned by episodic black-box policy search over movement-primitive parameters under a scalar survival-time reward~\cite{ploegerHighAccelerationReinforcement2021}.
The kinematic trajectory planner used here originated in our previous work~\cite{ploegerControllingCascadeKinematic2022}.
A 1g contact-switch assumption enables planning through dynamic contacts without explicit contact modeling, as demonstrated in simulation.
Extensions to non-uniform patterns, also in simulation, followed~\cite{andreuCascadeJugglingVanilla2024}.

Prior toss juggling has primarily focused on isolated subskills.
Catching has been studied with funnel-shaped end-effectors, fingered hands, and hybrid designs~\cite{baumlKinematicallyOptimalCatching2010,koberPlayingCatchJuggling2012,kimCatchingObjectsFlight2014}, with catch poses either intersecting predicted ballistic trajectories against a fixed plane~\cite{kizakiTwoBallJuggling2012} or optimizing within the reachable workspace~\cite{kimCatchingObjectsFlight2014}.
Throwing has relied on kinematic models~\cite{chenRobotThrowingTrajectory2019}, inverse dynamics~\cite{lombaiThrowingMotionGeneration2009}, hand-tuned movement primitives~\cite{sakaguchiStudyJugglingTasks1991,koberPlayingCatchJuggling2012}, and real-robot trial-and-error learning~\cite{aboafTasklevelRobotLearning1988a,koberLearningThrowingCatching2012,zengTossingBotLearningThrow2020}.
Full juggling systems range from bouncing automatons~\cite{shannonScientificAspectsJuggling1993,schaalOpenLoopStable1993} through planar fixed-actuator demonstrations~\cite{sakaguchiStudyJugglingTasks1991,burgetVisualfeedbackJugglerServo2010} to anthropomorphic arms with funnels or fingered hands~\cite{rileyRobotCatchingEngaging2002,kizakiTwoBallJuggling2012,okaBallJugglingRobot2017,ploegerHighAccelerationReinforcement2021}.
Robust five-ball juggling on robot arms remains open.

A recurring pattern for sample-efficient real-robot throwing is to keep an analytic ballistic model as the nominal release and learn only a corrective residual on top, supervised by the observed task error~\cite{aboafTasklevelRobotLearning1989,zengTossingBotLearningThrow2020}.
Aboaf et al.~\cite{aboafTasklevelRobotLearning1989} applied task-level corrections to single-arm juggling, the closest precedent.
The update rule that we deploy on hardware follows a similar task-level update structure, here scaled to multi-ball juggling across two arms and embedded in a structured comparison of feedback types and prior-specificity levels.
A different version of this pattern~\cite{zengTossingBotLearningThrow2020} uses a CNN conditioned on scene features to learn a \emph{prior} that generalizes to unseen objects on the first toss. We instead refine a \emph{posterior} on the residual from repeated attempts at a fixed configuration.

Three methodological antagonists frame our task-error formulation.
Gradient-free reinforcement learning approaches to robotic throwing~\cite{koberLearningThrowingCatching2012,ghadirzadehDeepPredictivePolicy2017,ploegerHighAccelerationReinforcement2021} consume only a scalar reward, discarding the directional structure of the task error and converging slowly on real robots.
Guided policy search~\cite{levineGuidedPolicySearch2013} restores directional information by computing cost gradients through a local dynamics model, but operates at the trajectory level and struggles with the contact dynamics central to juggling.
ILC~\cite{arimotoBetteringOperationRobots1984,bristowSurveyIterativeLearning2006} also uses directional feedback over repeated executions, but at the joint-tracking level rather than at the task level where the ball actually lands.
Task-error residual learning sits at the intersection: directional supervision lifted to the task outcome, without explicit integration through contact dynamics.

\clearpage
\section{System Overview}
\label{sec:system-overview}

The stack the residual learner sits on top of is simple, built for repeatability rather than accuracy,
with idealized assumptions about ball, robot, and contact dynamics throughout.
Each throw is open-loop with respect to the ball.
The kinematic trajectory planner (\cref{sec:trajectory-planning}) and joint-tracking controller (\cref{sec:ctrl-loop}) execute the throw without consulting visual feedback.
An external ball tracker observes the resulting flight, and the residual learner (\cref{sec:residual-learning}) consumes those observations to improve future throws by correcting the takeoff velocity.

\subsection{Kinematic Trajectory Planning}
\label{sec:trajectory-planning}

Joint trajectories $\q(t)$ are generated by solving the constrained velocity, acceleration, and jerk minimization problem
\begin{equation}
\begin{aligned}
\min_{\q(t)} \quad & \int_{0}^{T} \norm{\dq(t)}^2_{\mat{W}_{\dot{q}}} + \norm{\ddq(t)}^2_{\mat{W}_{\ddot{q}}} + \norm{\dddq(t)}^2_{\mat{W}_{\dddot{q}}} \dd t \\
\st \quad & \vec{h}(\x, \dx, \ddx) = \zeros, \\
& \vec{c}(\x, \dx, \ddx) \leq \zeros,
\end{aligned}
\label{eq:traj-opt}
\end{equation}
with equality constraints~$\vec{h}$ and inequality constraints $\vec{c}$ on the end-effector motion~$\x(t)$, and diagonal per-joint weights $\mat{W}_{\dot{q}}, \mat{W}_{\ddot{q}}, \mat{W}_{\dddot{q}}$.
We reuse the constraint structure from our previous work~\cite{ploegerControllingCascadeKinematic2022}.
The catch intercepts the desired ball trajectory at the planned touchdown; visual ball-tracking is available but does not enter the planning loop.
The throw matches the desired position, the takeoff velocity computed from an idealized parabolic ballistics model, and acceleration to gravity at takeoff.
At each contact switch we constrain the relative motion between the free-falling ball and the end-effector to avoid unintended contact, and during carry we bound the end-effector acceleration so that the ball remains seated.
The problem is solved by multiple shooting over piecewise-constant jerks in joint space, with $\vec{h}$ and $\vec{c}$ imposed through the time derivatives of the forward kinematics; we formulate it in CasADi and solve with IPOPT.

\subsection{Joint-Tracking Control}
\label{sec:ctrl-loop}

The arm is driven by a joint-space PD controller with rigid-body feedforward.
The feedforward torques account for link inertia, armature, and gravity only, taken from the manufacturer's CAD data without any system identification, and model no friction.
Gains are chosen overcritically damped; joint position and velocity are fused from joint and motor encoders.
The control loop is deliberately minimal: nothing in it is tuned or identified for this task, and any residual tracking error is left for the task-level learner to absorb.

\section{Residual Learning}
\label{sec:residual-learning}

A throw is specified by a target touchdown position $\ball_{\mathrm{TD}}^{*}$ and a planned flight time $t_{\mathrm{flight}}$.
With the flight time fixed at $t_{\mathrm{flight}}$, the parabolic ballistic map
\begin{equation}
    f(\vec{v}) = \vec{p}_0 + \vec{v}\, t_{\mathrm{flight}} + \tfrac{1}{2}\g\, t_{\mathrm{flight}}^2
    \label{eq:residual:f-def}
\end{equation}
sends a takeoff velocity $\vec{v}$ to the resulting touchdown position,
with $\vec{p}_0$ the takeoff position and $\g$ gravity.
Inverting the idealized map gives the nominal takeoff velocity command $\vec{v}_{\mathrm{TO}} = f\inv(\ball_{\mathrm{TD}}^{*})$.
The residual learner adds a per-throw correction $\vec{u}_n \in \R^3$, and the trajectory planner (\cref{sec:trajectory-planning}) takes the applied takeoff velocity $\vec{v}_{\mathrm{TO}} + \vec{u}_n$ as the release-state constraint that shapes the planned joint trajectory.
After each throw, ball tracking yields a task-error label $\vec{e}_n$ that measures how far the realized throw fell from the target, and the learner drives this error toward zero (\cref{sec:residual:label-generation}).
Updates are episodic, one observation per throw, on the real robot, and no learner is pretrained in simulation.

Each juggler runs several learner instances, one per throw index and arm in the transient phase and one per arm in the cyclic phase, all using the same method within a single experiment.
Later learners in each arm's chain are warm-started from the predecessor's current estimate.
In the cyclic phase the learner must commit to the next throw before the previous outcomes are observed.
We hold the candidate residual constant over subsequent throws until its feedback arrives, paying an $M\times$ per-update slowdown to keep the comparison across methods clean, where $M$ is set by flight time and tracking latency.
The remaining subsections describe how the learners turn the error sequence $\{\vec{e}_n\}$ into updates of $\vec{u}_n$.

\subsection{Task-Error Model and Learner Taxonomy}
\label{sec:residual:label-generation}
\label{sec:residual:axes}

Let $f_{\mathrm{true}}$ denote the true ballistic map at $t = t_{\mathrm{flight}}$.
Locally, around the nominal takeoff velocity, we approximate it as the idealized map plus an unknown constant offset
\begin{equation}
    f_{\mathrm{true}}(\vec{v}) \approx f(\vec{v}) + \vec{c},
    \label{eq:label:constant-offset}
\end{equation}
where $\vec{c}$ absorbs all model mismatch (drag, spin, motor calibration, contact dynamics) at a fixed throw configuration.
The residual-to-error-label map
\begin{equation}
    \emap(\vec{u}) := f\inv\!\big(f_{\mathrm{true}}(\vec{v}_{\mathrm{TO}} + \vec{u})\big) - \vec{v}_{\mathrm{TO}}
    \label{eq:label:g-def}
\end{equation}
composes the true ballistic map $f_{\mathrm{true}}$ with the idealized inverse and subtracts the nominal velocity.
On throw $n$ the ball lands at the measured touchdown $\ball_{\mathrm{TD},n}^{\mathrm{obs}} = f_{\mathrm{true}}(\vec{v}_{\mathrm{TO}} + \vec{u}_n)$, so the task error evaluates to
\begin{equation}
    \vec{e}_n = \emap(\vec{u}_n) = f\inv(\ball_{\mathrm{TD},n}^{\mathrm{obs}}) - \vec{v}_{\mathrm{TO}}.
    \label{eq:label:e-realized}
\end{equation}
We define $\emap$ to map into takeoff-velocity space.
By default, the observed touchdown position $\ball_{\mathrm{TD},n}^{\mathrm{obs}}$ comes from a least-squares parabolic fit to the late trajectory just before catching, projected to $t = t_{\mathrm{flight}}$ and back-projected through $f\inv$, so anchoring near touchdown absorbs ballistic-model error (chiefly air drag) into the label.
When that fit is corrupted by mid-air collisions or tracking dropouts, an early-trajectory fit reads the takeoff velocity directly, trading away ballistic-error correction for resilience.
Under the constant-offset approximation, the residual-to-error-label map
\begin{equation}
    \emap(\vec{u}) = \vec{u} + \vec{c}/t_{\mathrm{flight}} \qquad\text{with}\qquad \mat{J}_{\phi} = \pderiv{\emap}{\vec{u}} = \eye
    \label{eq:label:g-affine}
\end{equation}
is exactly affine, so the error Jacobian is the identity and the optimum residual sits at $\vec{u}^{*} = -\vec{c}/t_{\mathrm{flight}}$.
The identity $\mat{J}_{\phi} = \eye$ is the analytic prior the learners may adopt, while $\vec{u}^{*}$ is the unknown every learner must estimate from data.

\paragraph{Feedback type.}
\begin{table*}[b]
    \centering
    \caption{[\textsc{Learner Taxonomy}] Residual learners by feedback type (rows) and prior specificity (columns). \emph{None} makes no parametric assumption, \emph{Structural} fits the parameters of an assumed form, and \emph{Calibrated} fixes them to the analytic prior $\mat{J}_{\phi}=\eye$. The subsections below detail each cell.}
    \label{tab:residual:matrix-tier}
    \footnotesize
    \resizebox{\textwidth}{!}{%
    \begin{tabular}{|c|l|l|l|}
        \hline
        \diagbox{\textbf{Feedback}}{\textbf{Prior}} & \multicolumn{1}{c|}{\textbf{None}} & \multicolumn{1}{c|}{\textbf{Structural}} & \multicolumn{1}{c|}{\textbf{Calibrated ($\mat{J}_{\phi}=\eye$)}} \\
        \hline
        \textbf{Directional} ($\vec{e}$)        & Per-axis $(1{+}1)$-ES   & MLE Jacobian        & Fixed Jacobian       \\
                                                &                         & Composite BO        & MAP Jacobian         \\
                                                &                         &                     & Composite BO         \\
        \hline
        \textbf{Norm} ($\norm{\vec{e}}$)        & CMA-ES, REPS            & BO + cone    & BO + cone     \\
                                                &                         &                     &                      \\
        \hline
        \textbf{Squared} ($\norm{\vec{e}}^2$)   & CMA-ES, REPS            & BO + paraboloid & BO + paraboloid \\
                                                &                         &                     &                      \\
        \hline
    \end{tabular}%
    }
\end{table*}

A central question of this work is how much the directional structure of the task error is worth.
We compare learners that consume the full error vector $\vec{e}_n$ against learners that see only a scalar summary of the error, either $\norm{\vec{e}_n}$ or $\norm{\vec{e}_n}^2$,
reformulating the root-finding problem as a cost minimization.
Under the constant-offset model, all three feedback types
\begin{align}
    \vec{e}          &= \mat{J}\,(\vec{u} - \vec{u}^{*}), & &\text{(affine)} \label{eq:label:form-affine} \\
    \norm{\vec{e}}   &= \norm{\mat{J}\,(\vec{u} - \vec{u}^{*})}, & &\text{(cone)} \label{eq:label:form-cone} \\
    \norm{\vec{e}}^2 &= (\vec{u} - \vec{u}^{*})\trans \mat{J}\trans \mat{J}\, (\vec{u} - \vec{u}^{*}) & &\text{(paraboloid)} \label{eq:label:form-paraboloid}
\end{align}
are parameterized by the task-error Jacobian $\mat{J}$ forming the rows of \cref{tab:residual:matrix-tier}.

\paragraph{Prior specificity.}
Reading the columns of \cref{tab:residual:matrix-tier} left to right, a learner commits to progressively more of this parametric form.
\emph{None} assumes no parametric form and updates from stochastic search on the raw feedback.
\emph{Structural} fixes the functional form (affine, cone, or paraboloid) and fits its parameters, including the Jacobian $\mat{J}$, from data.
\emph{Calibrated} fixes both the form and its values, setting $\mat{J} = \mat{J}_{\phi} = \eye$ from \cref{eq:label:g-affine} and leaving only the optimum location $\vec{u}^{*}$ free.
The remaining subsections describe a representative method for each cell.

\subsection{Jacobian-Based Newton Updates}
\label{sec:residual:fitted-jacobian}

The three directional Newton methods of \cref{tab:residual:matrix-tier} treat residual learning as root-finding on the task error $\vec{e}(\vec{u})$: linearize at the current command, $\vec{e}(\vec{u}) \approx \vec{e}_n + \hat{\mat{J}}_n (\vec{u} - \vec{u}_n)$, and step toward the command that zeros the linearization.
With a local Jacobian estimate $\hat{\mat{J}}_n$, the damped update is
\begin{equation}
    \vec{u}_{n+1} = \vec{u}_n - \alpha_n\, \hat{\mat{J}}_n^{+}\, \vec{e}_n,
    \label{eq:residual:nr-step}
\end{equation}
where $\hat{\mat{J}}_n^{+}$ is the Moore--Penrose pseudo-inverse and $\alpha_n$ is an exponential damping schedule that trades early speed against noise-driven thrashing near the error floor.
The cells differ only in $\hat{\mat{J}}_n$.
The data-driven flavors relate each past sample's command deviation $\Delta\vec{u}_k$ to its error deviation $\Delta\vec{e}_k$ from the current operating point,
by kernel-weighted local least squares
\begin{equation}
    \hat{\mat{J}}_n = \argmin_{\mat{J}}\; \sum_k \kappa(\Delta\vec{u}_k)\, \norm{\Delta\vec{e}_k - \mat{J}\,\Delta\vec{u}_k}^2 + \lambda \norm{\mat{J} - \mat{J}_0}_F^2,
    \label{eq:residual:jacobian-fit}
\end{equation}
with a Gaussian kernel $\kappa$.
The three Jacobian-based methods in \cref{tab:residual:matrix-tier} differ only in their choice of $(\mat{J}_0, \lambda)$.
The \emph{MLE Jacobian} (Structural; maximum-likelihood estimate, MLE) fits $\hat{\mat{J}}_n$ from data alone, retaining only a small $\lambda$ toward $\mat{J}_0 = \zeros$ for numerical stability.
The \emph{MAP Jacobian} (Calibrated; maximum a posteriori, MAP) sets the prior to the analytic value $\mat{J}_0 = \mat{J}_{\phi} = \eye$ from \cref{eq:label:g-affine} with finite $\lambda$, so it steps along the prior where data is absent and approaches the MLE fit as data accumulates.
The \emph{Fixed Jacobian} (Calibrated) is its $\lambda \to \infty$ limit: data is ignored and $\hat{\mat{J}}_n = \mat{J}_{\phi} = \eye$ throughout, collapsing \cref{eq:residual:nr-step} to $\vec{u}_{n+1} = \vec{u}_n - \alpha_n \vec{e}_n$.

Two practical safeguards apply on top of the Newton step for the data-driven flavors.
When $\hat{\mat{J}}_n$ is poorly conditioned (data sparse or correlated in $\vec{u}$-space), the step is augmented with a stochastic perturbation
\begin{equation}
    \Delta\vec{u}_n \;\leftarrow\; \Delta\vec{u}_n + \abs{\xi}\, \vec{d}_n, \qquad \xi \sim \mathcal{N}(0,\, \sigma_{\mathrm{explore}}^2),
    \label{eq:residual:nr-explore}
\end{equation}
along the smallest-singular-vector direction $\vec{d}_n$ of $\hat{\mat{J}}_n$, exploring the residual direction the data has failed to constrain.
The damped step is then accepted only if the task-error norm decreases below the most recently accepted error; the new sample is added to the regression of \cref{eq:residual:jacobian-fit} either way, so a rejected step still shifts $\hat{\mat{J}}_{n+1}$ and influences the next proposed command.

\subsection{Bayesian Optimization}
\label{sec:residual:bayesian-optimization}

The Bayesian Optimization (BO) methods of \cref{tab:residual:matrix-tier} model the objective with a Gaussian-process (GP) surrogate, a Mat\'ern-$5/2$ kernel with hyperparameters fit per attempt and the posterior refreshed at every query, and choose the next command as the lower-confidence-bound minimizer
\begin{equation}
    \vec{u}_{n+1} = \argmin_{\vec{u}}\; \mu_n(\vec{u}) - \beta\, \sigma_n(\vec{u}),
    \label{eq:residual:bo-lcb}
\end{equation}
with posterior mean $\mu_n$, standard deviation $\sigma_n$, and exploration weight $\beta$, implemented using BoTorch~\cite{balandatBoTorchFrameworkEfficient2020}.
The scalar methods regress the cost, with the analytic shape from \cref{sec:residual:axes} as the GP prior mean: the cone \mbox{$m(\vec{u}) = \norm{\mat{J}(\vec{u} - \vec{u}^{*})}$} for the norm cost, the paraboloid \mbox{$m(\vec{u}) = (\vec{u} - \vec{u}^{*})\trans \mat{Q}\, (\vec{u} - \vec{u}^{*})$} with curvature $\mat{Q} = \mat{J}\trans \mat{J}$ for the squared norm.

The directional (composite) methods instead place a multi-output GP on $\vec{e}(\vec{u})$ itself with the affine mean $\vec{m}(\vec{u}) = \mat{J}(\vec{u} - \vec{u}^{*})$.
Since the analytic Jacobian is diagonal, the three channels are modeled independently.
The resulting Gaussian posterior $\vec{e}(\vec{u}) \sim \mathcal{N}(\vec{\mu}_n(\vec{u}), \Sigma_n(\vec{u}))$ is projected onto the scalar cost $\norm{\vec{e}}^2$ in closed form,
\begin{equation}
    \E\!\left[\norm{\vec{e}}^2\right] = \norm{\vec{\mu}_n}^2 + \trace(\Sigma_n), \qquad
    \mathrm{Var}\!\left[\norm{\vec{e}}^2\right] = 2\, \trace(\Sigma_n^2) + 4\, \vec{\mu}_n\trans \Sigma_n\, \vec{\mu}_n,
    \label{eq:residual:vbo-moments}
\end{equation}
and the projected mean and standard deviation enter the same lower-confidence-bound acquisition of \cref{eq:residual:bo-lcb} in place of $\mu_n, \sigma_n$.

All \emph{Structural} methods fit the mean's parameter ($\mat{J}$ or $\mat{Q}$) from data by marginal likelihood, while \emph{Calibrated} methods fix it at $\mat{J} = \mat{J}_{\phi} = \eye$ from \cref{eq:label:g-affine}.
The scalar BO variants additionally fit the optimum location $\vec{u}^{*}$ of the cost from data, whereas Composite BO has no such parameter, since it models the task error $\vec{e}(\vec{u})$ rather than the cost.

Calibrated variants take $\vec{u}_0 = \zeros$ as the first command, while Structural variants instead seed a small Sobol pool around $\zeros$ to bootstrap the parameter fit.
Combined with the analytic mean from \cref{sec:residual:axes}, the GP performs a local search around $\vec{u} = \zeros$ rather than the global search a flat-mean GP would run on the full residual space.
Composite BO relates conceptually to the Jacobian flavors of \cref{sec:residual:fitted-jacobian}, with the difference that Composite BO trades substantially more compute for uncertainty estimates over $\vec{e}$, where the Jacobian flavors settle for a point estimate of $\hat{\mat{J}}_n$.

\subsection{Stochastic Search}
\label{sec:residual:stochastic-search}

All methods in the \emph{None} column of \cref{tab:residual:matrix-tier} avoid parametric assumptions and explore by random perturbation of the command, without ever fitting a Jacobian or surrogate, the baseline against which the \emph{Structural} and \emph{Calibrated} priors are measured.
For the two scalar cells we run two black-box optimizers on the cost $\norm{\vec{e}}$ and $\norm{\vec{e}}^2$: CMA-ES with population-based covariance adaptation for noise robustness~\cite{hansenCompletelyDerandomizedSelfAdaptation2001}, and episodic REPS, which maximizes the negative cost under a KL-bounded distribution update~\cite{petersRelativeEntropyPolicy2010}.
These scalar searches are the scalar-reward reinforcement-learning baseline, expected to be the least sample-efficient of the methods we compare.

As a directional variant, we run an independent $(1{+}1)$-ES per axis, deciding each step's acceptance and step-size adaptation on $\abs{e_i}$ rather than on the norm.
This exploits the directional feedback with the least commitment to a parametric form, by only assuming that each error channel depends primarily on its own residual dimension.

\clearpage
\section{Experiments}
\label{sec:experiments}

The experiments establish both which residual formulation to use and how little the surrounding stack must be trusted.
We select a learner in simulation, demonstrate it juggling on the real robot, and then degrade the analytic prior and the tracking controller in turn to map its operating range.

\subsection{Experimental Setup}
\label{sec:experimental-setup}
We evaluate the residual learners both on a real two-arm robot and in a MuJoCo simulation of the same system, sweeping the full method matrix in simulation.
The hardware is a pair of \mbox{4-DOF} Barrett WAM~3.0 arms with dual joint and motor encoders, run at \SI{160}{\volt} with custom cooling at a \SI{1}{\kilo\hertz} control rate.
Each arm carries a 3D-printed, funnel-shaped end-effector at the wrist for catching and throwing.
Balls are tracked with an OptiTrack motion-capture system using per-ball Kalman filters with scheduled dynamics-model switches and Hungarian data association.
Simulation and hardware share the planner (\cref{sec:trajectory-planning}) and controller (\cref{sec:ctrl-loop}); only the physics is replaced by MuJoCo.
The unmodeled ball payload and impacts add uncontrolled joint-tracking error.
We perturb the ball velocity after takeoff to reflect the finite repeatability of the real system.
We randomize the plant by scaling each arm's link masses and joint armatures by $\pm\SI{30}{\percent}$, seeded per run, so all methods face the same plants.
The converged residual magnitude in simulation matches the real-robot value within \SI{20}{\percent}.

\subsection{Method Comparison}
\label{sec:exp:method-comparison}

\begin{figure}[t]
    \centering
    \includegraphics[width=\linewidth]{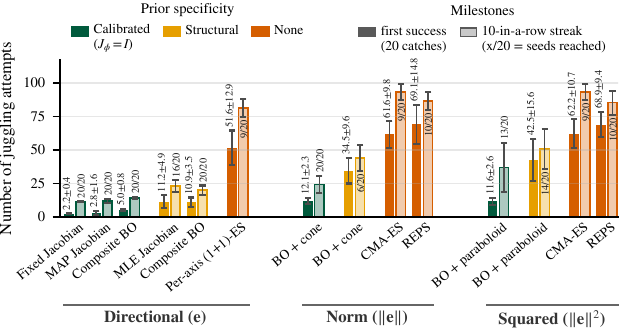}
    \caption{[\textsc{Method Comparison}] Fast, reliable convergence needs both directional feedback and an informative prior. Evaluated in simulation at five-ball cascade over the feedback~$\times$~prior-specificity matrix of \cref{tab:residual:matrix-tier}, grouped by feedback type and colored by prior specificity. Bars give attempts to the first success over 20 throws (solid) and to a ten-in-a-row streak (light; reached by $X/20$ seeds). Values are mean $\pm$ s.d.\ over 20 seeds spanning 10 randomized plants.}
    \label{fig:exp:method-comparison:sim-attempts}
\end{figure}

We sweep the full $3\times 3$ method matrix of \cref{tab:residual:matrix-tier} in simulation at five-ball cascade, with 20 seeds per cell, measuring the number of attempts to the first success and to a ten-in-a-row streak (\cref{fig:exp:method-comparison:sim-attempts}), where a simulated attempt succeeds once it completes 20 throws.
Sample efficiency degrades almost monotonically along both axes, and the two effects compound: directional feedback beats scalar feedback, whether norm or squared norm, and a more specific prior beats a vaguer one (calibrated $\to$ structural $\to$ none).
Squaring the cost changes how the search behaves near the optimum.
Around the minimum the squared cost is nearly flat, so the surrogate barely distinguishes the optimum from its neighborhood and keeps exploring there, whereas the cone's constant-magnitude slope makes every step away from the optimum visibly costly.
In the calibrated cells both means reach a first success equally fast ($11.6$ and $12.1$ attempts), but the paraboloid's continued exploration costs consistency: it holds a ten-in-a-row streak on only 13 of 20 seeds and needs $36.9$ attempts on those that do, against the cone's $24.2$ on all 20.
The only cell that is both fast and fully reliable is directional feedback paired with a calibrated prior, where the Fixed Jacobian, MAP Jacobian, and Composite BO learners each converge within a handful of attempts and reach a streak on all 20 seeds.
Among these the Fixed Jacobian Newton update is also the simplest, with no data-driven Jacobian fit and no surrogate model to optimize, so the best-performing learner is also the cheapest to run.
We therefore adopt it as the canonical learner for all of the real-robot experiments that follow.

\begin{table}[b!]
    \centering
    \caption{[\textsc{Real-Robot Juggling}] The Fixed Jacobian residual converges near-immediately, bringing all three patterns to a reliable juggle within two attempts. Each pattern is run over six seeds of ten attempts each, and every attempt executes up to 120 throws, about \SI{30}{\second} at four and five balls, counting as a success only when the pattern completes them all. The converged residual command norm $\norm{\vec{u}}$ and the noise floor $\sigma_{\norm{\vec{e}}}$ are given in \si{\meter\per\second}.}
    \label{tab:exp:real-robot}
    \footnotesize
    \setlength{\tabcolsep}{6pt}
    \begin{tabular}{lcccc}
        \hline
        Pattern & 1st succ. & 1st 3-in-a-row & Residual $\norm{\vec{u}}$ & Noise floor $\sigma_{\norm{\vec{e}}}$ \\
        \hline
        3-ball cascade  & $1.2\pm0.4$ & $3.2\pm0.4$ & $0.19\pm0.02$ & $0.016$ \\
        4-ball fountain & $1.8\pm0.4$ & $3.8\pm0.4$ & $0.22\pm0.02$ & $0.019$ \\
        5-ball cascade  & $2.2\pm0.4$ & $4.2\pm0.4$ & $0.23\pm0.01$ & $0.022$ \\
        \hline
    \end{tabular}
\end{table}

\subsection{Real-Robot Juggling}
\label{sec:exp:real-robot}

We then deploy the Fixed Jacobian residual on the real robot across all three patterns (\cref{tab:exp:real-robot}).
Convergence is near-immediate: the three-ball cascade typically succeeds on the first attempt and the four- and five-ball patterns by the second, after which all three hold a steady streak of completed runs.
Each pattern is evaluated over six seeds of ten attempts, with every attempt running up to 120 throws, about \SI{30}{\second} of continuous juggling at four and five balls, and counting as a success only once it completes all 120 throws.

The residual command norm $\norm{\vec{u}}$ is the magnitude of the velocity correction the learner settles on, around \SI{0.2}{\meter\per\second} across patterns.
The noise floor $\sigma_{\norm{\vec{e}}}$ is the irreducible run-to-run variability of the real system.
Propagated ballistically through the \SI{1.00}{\second} flight time, the five-ball noise floor corresponds to a steady-state landing-position scatter of about \SI{22}{\milli\meter} at the catch plane.

\begin{figure}[t]
    \centering
    \includegraphics[width=\linewidth]{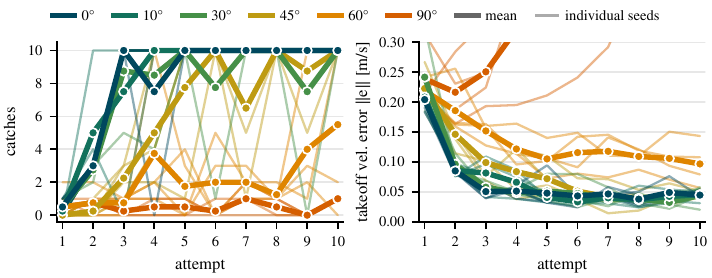}
    \caption{[\textsc{Prior-Quality Ablation}] Residual learning tolerates substantial misalignment of the analytic prior. The analytic Jacobian is rotated about four fixed random axes, one per seed, $\mat{J}' = \mat{R}\,\mat{J}$, by $0$--$\SI{90}{\degree}$ on the real robot with the Fixed Jacobian learner at five-ball cascade. The left panel shows the catches per attempt, capped at ten; the right panel shows the transient task-error norm $\norm{\vec{e}}$. Each setting is run over four seeds of ten attempts each.}
    \label{fig:exp:prior-quality:rotation}
\end{figure}

\subsection{Prior-Quality Ablation}
\label{sec:exp:prior-quality}

Both real-robot ablations share tighter joint limits than the headline runs of \cref{sec:exp:real-robot}, and score success over ten throws rather than 120.
Under degraded tracking the arms can overswing far enough to tip the end-effector forward and let the balls roll out, so we constrain the joints in the stack-accuracy sweep and, for comparability, apply the same limits to the prior-quality ablation.
The tighter limits require a larger residual and shift convergence from the second attempt to the third.

The Fixed Jacobian learner starts from an identity prior.
To probe how much prior quality matters, we deliberately degraded it: the analytic Jacobian is rotated about a random axis by $0$ to \SI{90}{\degree}, and the residual must converge through the misalignment (\cref{fig:exp:prior-quality:rotation}).
The learner absorbs substantial misalignment: rotations up to \SI{30}{\degree} leave convergence speed unchanged, and even a \SI{60}{\degree} rotation still converges.
Slower, but still faster and more reliably than any scalar or stochastic-search method in simulation (\cref{fig:exp:method-comparison:sim-attempts}).
Directional feedback explains why.
As long as the rotated Jacobian keeps a positive projection onto the true descent direction, each Newton step still reduces the error.
Only the degenerate \SI{90}{\degree} rotation fails, where that projection vanishes and the prior no longer points toward lower error.

\subsection{Stack-Accuracy Ablation}
\label{sec:exp:stack-accuracy}

\begin{figure}[t]
    \centering
    \includegraphics[width=\linewidth]{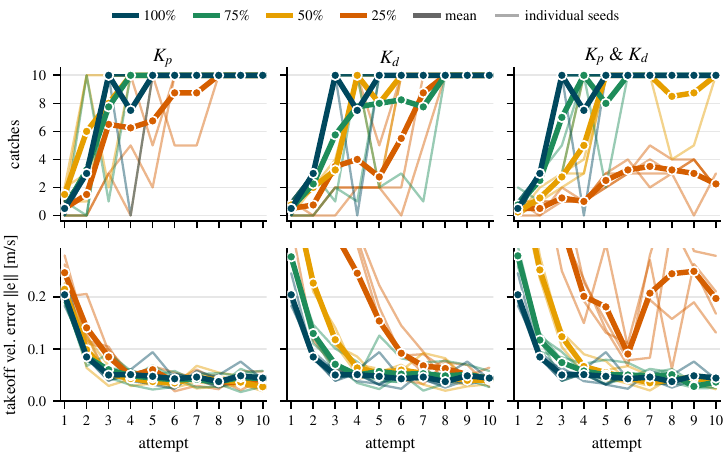}
    \caption{[\textsc{Stack-Accuracy Ablation}] Final task performance is invariant to controller-gain quality, with weaker gains affecting only the convergence speed. The robot runs the Fixed Jacobian learner at five-ball cascade. Each column weakens a different gain set from \SI{100}{\percent} down to \SI{25}{\percent} of nominal. The top row reports catches per attempt and the bottom row the transient task-error norm $\norm{\vec{e}}$. Each setting is run over four seeds of ten attempts each.}
    \label{fig:exp:stack-accuracy:gains}
\end{figure}

The complementary question is how much the quality of the underlying controller matters.
We scaled the nominal PD gains, $\vec{K}_p = (400, 400, 200, 200)\,\si{\newton\meter\per\radian}$ and $\vec{K}_d = (40, 20, 15, 5)\,\si{\newton\meter\second\per\radian}$ from base to wrist, down to as little as \SI{25}{\percent} of their default values, sweeping the proportional gains, the derivative gains, and both together (\cref{fig:exp:stack-accuracy:gains}).
Final task performance is unchanged across the entire sweep.
Every setting reaches the same task-error floor and the same reliable juggle, and only the transient differs, with weaker gains needing more attempts to converge.
The reason is repeatability rather than accuracy.
A softer controller tracks less accurately, but it does so consistently, making the same error on every throw; the residual simply absorbs that fixed offset, and the throws keep landing where they should.
Gain quality therefore sets how large a correction the residual must learn and how quickly it gets there, not the final result.
This holds as long as the executed motion stays coupled to the commanded one.
Under more extreme gain reductions the coupling weakens and the residual loses its grip on the plant.

\clearpage
\section{Conclusions}
\label{sec:conclusions}
We presented task-error residual learning on real anthropomorphic arms juggling three-, four-, and five-ball patterns.
The unlearned stack uses a 1g contact-switch planner, a parabolic ballistic predictor, and a soft PD controller with rigid-body feedforward, rough by design.
Across all three patterns the residual learner converges by the second attempt and holds steady.
The empirical method comparison shows that directional feedback and an informative prior are both necessary: neither suffices alone, and only their combination converges in a handful of attempts.
The simplest method we tested, a Fixed Jacobian Newton update with an identity prior, is also the most effective and reliable on hardware.

A full evaluation of all methods on the real robot would take prohibitively much wall time, so the $3\times 3$ matrix is swept only in simulation.
The real-robot ablations use the Fixed Jacobian throughout.
The prior-quality ablation shows the learner tolerates substantial rotation of the analytic Jacobian.
The stack-accuracy sweep shows final task error invariant to PD gain scaling from \SI{100}{\percent} down to \SI{25}{\percent}, with only convergence speed varying.
This invariance holds for as long as the executed motion remains coupled to the commanded one.
Under strongly reduced gains the coupling weakens and the residual approach degrades.
The scalar-feedback critique applies wherever sample efficiency matters.
In simulation that cost is bounded by compute, but the bottleneck for transferring sim-trained policies to highly dynamic throwing tasks is the sim-to-real gap.

Our work hints at a way to narrow that gap: transferring a robust \emph{adaptation process} alongside the nominal policy, rather than the policy alone.
The priors used here, such as $\mat{J}_{\phi} = \eye$, are simple enough to be extracted from a simulator autonomously.
Our prior-rotation results suggest that simple local adaptation processes transfer well across modest model mismatch, which is exactly the regime a sim-to-real gap introduces.
Two other directions follow.
Contextual residual learning would propagate catch-side outcomes into the next throw's residual rather than absorbing them into a static vector.
Non-uniform juggling patterns require no methodological changes but additional task-decomposition machinery to handle multiple learner roles per throw.

\begingroup
\sloppy   %
\subsubsection*{Acknowledgement.}
This work was supported by the Deutsche Forschungsgemeinschaft (German Research
Foundation, DFG) under Germany's Excellence Strategy (EXC 3066/1 ``The Adaptive
Mind'', Project No.~533717223), and by the Hessian Ministry of Higher Education,
Research, Science and the Arts and its LOEWE research priority program
``WhiteBox'' under grant
LOEWE/\allowbreak 2/\allowbreak 13/\allowbreak 519/\allowbreak 03/\allowbreak 06.001(0010)/\allowbreak 77.
The authors used Claude (Anthropic) for language editing and proofreading.
All technical content is the authors' own, and the authors take full
responsibility for the final text.
\endgroup

\clearpage

\bibliography{reference/references_hybrid}
\bibliographystyle{splncs03}

\end{document}